# Design of Mobile Manipulator for Fire Extinguisher Testing. Part II: Design and Simulation

Thai Nguyen Chau[1], Xuan Quang Ngo[1], Van Tu Duong[1,2,3], Trong Trung Nguyen[4], Huy Hung Nguyen[5] and Tan Tien Nguyen[1,2,3(✉)]

[1] National Key Laboratory of Digital Control and System Engineering (DCSELab), Ho Chi Minh City University of Technology (HCMUT), 268 Ly Thuong Kiet Street, District 10, Ho Chi Minh City, Vietnam
[2] Faculty of Mechanical Engineering, Ho Chi Minh City University of Technology (HCMUT), 268 Ly Thuong Kiet, District 10, Ho Chi Minh City, Vietnam
[3] Vietnam National University Ho Chi Minh City, Linh Trung Ward, Thu Duc District, Ho Chi Minh City, Vietnam
[4] Ho Chi Minh City University of Transport, Vietnam
[5] Faculty of Electronics and Telecommunication, Saigon University, Vietnam
`nttien@hcmut.edu.vn`

**Abstract.** All flames are extinguished as early as possible, or fire services have to deal with major conflagrations. This leads to the fact that the quality of fire extinguishers has become a very sensitive and important issue in firefighting. Inspired by the development of automatic fire fighting systems, this paper presents a mobile manipulator to evaluate the power of fire extinguishers, which is designed according to the standard of fire extinguishers named as ISO 7165:2009 and ISO 11601:2008. A detailed discussion on key specifications solutions and mechanical design of the chassis of the mobile manipulator has been presented in Part I: Key Specifications and Conceptual Design. The focus of this part is on the rest of the mechanical design and controller design of the mobile manipulator.

## 1    Manipulator Design

### 1.1    Forward Kinematics

The manipulator design is inherited from well-known industrial manipulators such as Motorman MPL500II of Yaskawa [1], which can implement the same task as humans. The design of a mobile manipulator is required to meet the key specifications. Based on the requirement for the fire tests class A and B, design parameters for the mobile manipulator are assumed in Table 1 and Table 2 with the coordinate system as shown in Fig. 1. The kinematic diagram is described in Part I: Key Specifications and Conceptual Design.

**Table 1.** General specifications of the fire palletizing manipulator

| | Fire testing manipulator parameters | | |
|---|---|---|---|
| | D.O.F: 4 | Load capacity: 20$kg$ | Max reach: 2255 $mm$ |
| Operation range of joints (Zero angle is at horizontal axis of the corresponding coordinate) | Joint A | $\pm 180^o$ | Rotate around z-axis of Coord0 |
| | Joint B | $0^o$ to $+135^o$ | Rotate around z-axis of Coord1 |
| | Joint D | $-120^o$ to $+15.5^o$ | Rotate around z-axis of Coord2 |
| | Joint G | $-90^o$ to $0^o$ | Rotate around z-axis of Coord4 |

**Table 2.** Denavit-Hartenberg (D-H) parameters of the palletizing manipulator

| Frame No. | Coord1 | Coord2 | Coord3 | Coord4 | Coord5 |
|---|---|---|---|---|---|
| $a_i$ (mm) | 100 | 1000 | 1700 | 100 | 80 |
| $\alpha_i$ (deg) | 90 | 0 | 0 | 0 | 0 |
| $d_i$ (mm) | 185 | 0 | 0 | 0 | 0 |
| $\theta_i$ (deg) | $\theta_1$ | $\theta_2$ | $\theta_3$ | 0 | $\theta_4$ |

In Fig. 1, the coordinate frames of linkages using D-H convention are established. D-H parameters are listed in Table 2 with frame {0} fixed and the end-effector point P.

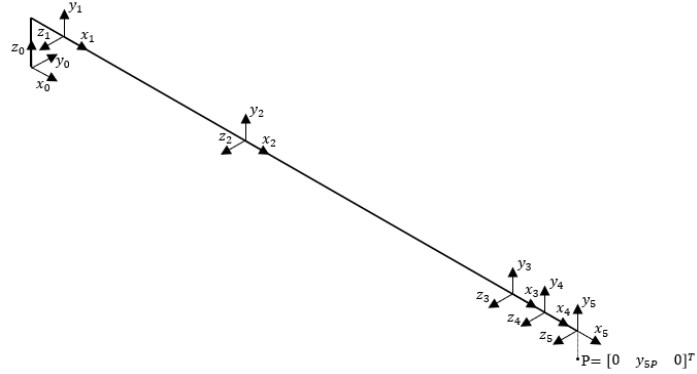

**Fig. 1.** Manipulator coordinate system

The D-H transformation matrix describes Coord($i$) relative to Coord($i-1$) presented in reference [2]. The position of the end-effector $P_5 = [0 \quad y_{5P} \quad 0 \quad 1]^T$ which belongs to Coord5, is determined in the global coordinate:

$$P_0 = T_5^0 P_5 = T_4^0 T_5^4 P_5 = \begin{bmatrix} c_1(a_1 + a_2 c_2 + a_3 c_3 + a_4 + a_5 c_4 - y_{5P} c_1 s_4) \\ s_1(a_1 + a_4 + a_5 s_4 + a_3 c_{23} - a_2 s_2 - y_{5P} s_1 s_4) \\ d_1 + a_2 s_2 + a_3 s_3 + a_5 s_4 + y_{5P} c_4 \\ 1 \end{bmatrix} \quad (1)$$

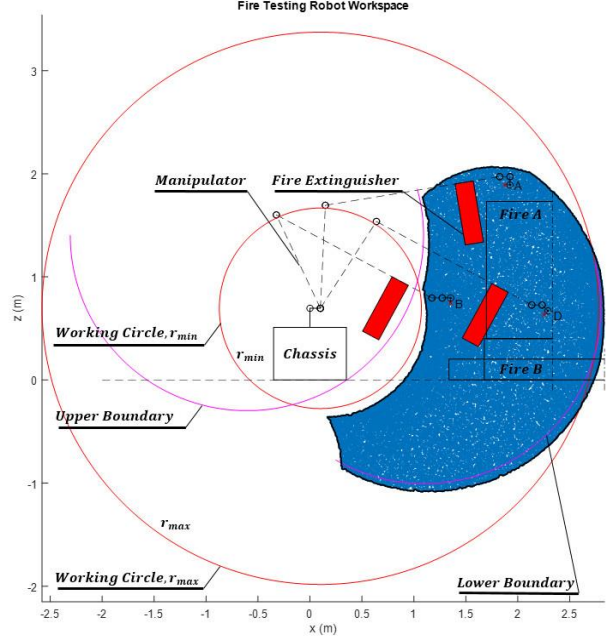

**Fig. 2.** Workspace of mobile manipulator

From the parameters listed in Table 1, Table 2, and Eq. (1), the workspace analysis of the manipulator as shown in Fig. 2 is conducted within the following constraints. The minimal and maximal angles formed by the second and third linkage are $30^o$ and $165^o$ respectively. Therefore, the minimum and maximum working radius are calculated by applying trigonometric functions: $r_{min} = 972 \ mm$ and $r_{max} = 2678 \ mm$. The workspace is defined by combination of working circles with maximum radius and minimum radius. Upper boundary is when $\theta_2 = 135^o$; $-120^o \leq \theta_3 \leq 15.5^o$, and lower boundary is when $\theta_2 = 0^o$; $-120^o \leq \theta_3 \leq 15.5^o$.

The simulation of manipulator workspace described in Fig. 2 shows that all the working positions, namely point A, B and D mentioned in the key specification of Part I, are covered in the workspace of the manipulator. This results in the fact that the initialization data for the manipulator fulfills the key specifications.

## 1.2 Inverse Kinematics

Inverse kinematics determines the angle of each joint when the end-effector pose $P = [x_P \quad y_P \quad z_P \quad \phi]^T$, $\phi$ is the end-effector orientation, is available. To compute the angles for joints 2, 3 and 4, $x-$coordinate and $y-$coordinate is combined into $r-$coordinate as shown in Fig. 3 using the Pythagorean theorem, $r_P = \sqrt{x_P^2 + y_P^2}$.

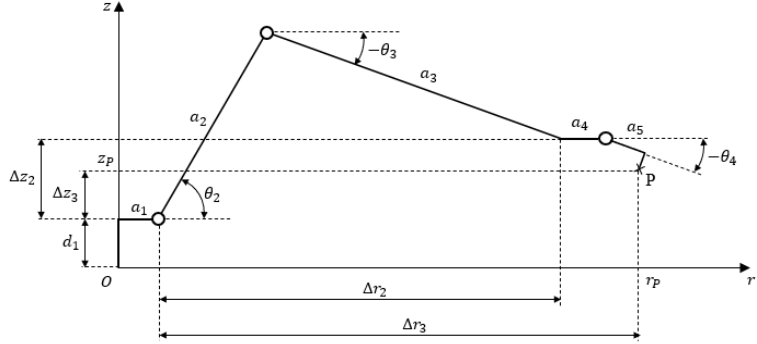

**Fig. 3.** $r - z$ coordinate system for the designed manipulator

The solutions of the inverse kinematics of the manipulator are given as:

$$\theta_1 = \tan^{-1}(y_P, x_P); \quad \theta_4 = -\phi$$

$$\theta_2 = \cos^{-1}\left(\frac{\Delta r_2^2 + \Delta z_2^2 + a_2^2 - a_3^2}{2a_2\sqrt{\Delta r_2^2 + \Delta z_2^2}}\right) + \tan^{-1}\frac{\Delta z_2}{\Delta r_2} \qquad (2)$$

$$\theta_3 = -\cos^{-1}\left(\frac{\Delta r_2^2 + \Delta z_2^2 + a_3^2 - a_2^2}{2a_3\sqrt{\Delta r_2^2 + \Delta z_2^2}}\right) + \tan^{-1}\frac{\Delta z_2}{\Delta r_2}$$

where, $\Delta r_3 = r_P - a_1$; $\Delta z_3 = z_P - d_1$; $\Delta r_2 = \Delta r_3 + y_{5P}\sin\theta_4 - a_4 - a_5\cos\theta_4$; $\Delta z_2 = \Delta z_3 - a_5\sin\theta_4 - y_{5P}\cos\theta_4 + d_4$.

### 1.3 Dynamic Modeling

The dynamic equation for the manipulator with the torque $Q$ are obtained:

$$M(q)\ddot{q} + C(q,\dot{q})\dot{q} + G(q) = Q \qquad (3)$$

where, $q = [\theta_1 \ \theta_2 \ \theta_3 \ \theta_4]^T$ are the joint angles; $M(q)$ is the symmetric inertia matrix, $C(q,\dot{q})$ is the centrifugal/ Coriolis matrix; $G(q)$ is the gravitational matrix.

In this paper, Lagrange law which provides only the required differential equations that determines the actuators' force and torque is used [3]. Applying the Lagrange law, the motion of manipulator can be described as:

$$\frac{d}{dt}\left(\frac{\partial \mathcal{L}}{\partial \dot{\theta}_i}\right) - \frac{\partial \mathcal{L}}{\partial \theta_i} = Q_i \qquad (4)$$

where, $\mathcal{L} = K - V$ is the difference between kinetic $K$ and potential $V$ energies. $K$ and $V$ are described clearly in reference [3]. Then the inertia matrix $M$ and the gravitational $G$ are retrieved, respectively. The centrifugal/Coriolis term $C(\dot{q},q)$ is computed using Christoffel symbols referring to [4].

## 2 Control Algorithm

The mobile manipulator can be operated when initial parameters such as the type, mass and discharge time of fire extinguishers and the class of fire tests are entered. Then, the mobile manipulator moves to "Home" position where a simulated fire is located. The process of putting out the fire test includes two stages as described in Fig. 4.

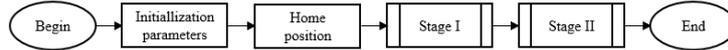

**Fig. 4.** Main control algorithm flowchart for the fire extinguisher testing system

The manipulator sticks to the known plan depending on the class and power of the fire test when comes to Stage I. This step leads to the fact that the fire test is covered completely with the substance used in the fire extinguisher. Fig. 5 and Fig. 6 describe the program to extinguish the class 20A fire test. The chassis controller drives the chassis for the trajectory II with five important points $Ki$, as shown in Fig. 14. This trajectory not only allows the mobile manipulator to fulfill the requirements of the key specification but also keeps the operation smooth and continuing. After the chassis moves to point $K1$, the manipulator controller drives the manipulator to perform the trajectory I, as shown in Fig. 11. The manipulator operation is required to complete before the chassis arrives point $K2$. The operation is repeated until the chassis reaches point $K4$. In the later part of Stage I, the chassis go to point $K5$ with the manipulator configuration reaching point A as shown in Fig. 2 to spray the agent from the top.

In Stage II, a thermal camera detects remaining small flames. Hence, these flames are handled sequentially with once movement around the fire test. Otherwise, the chassis goes to "End" position as described in Fig. 6 to finish the testing process.

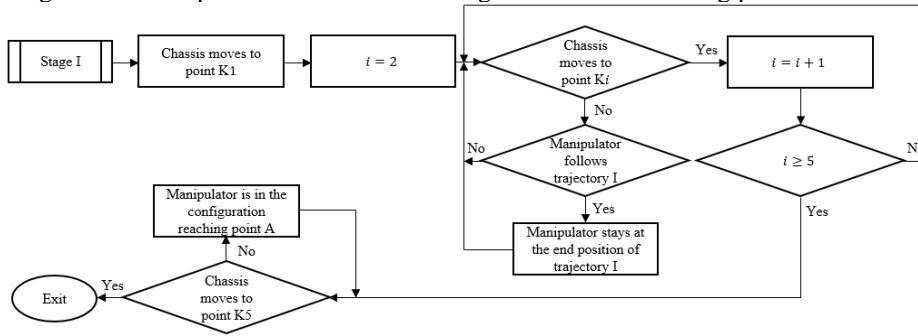

**Fig. 5.** Stage I in the process of putting out class A fire tests

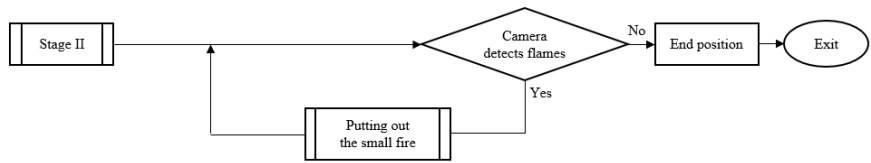

**Fig. 6.** Stage II in the process of putting out class A fire tests

## 2.1 Algorithm I

After finishing Stage I, the chassis is at point $K5$. Algorithm I consists of two main parts as shown in Fig. 7.

Firstly, Algorithm I.1 as shown in Fig. 8 finds out the trajectory II's positions $Ci$ where the mobile manipulator needs to actuate the fire extinguisher to put out flames $Fi$ as shown in Fig. 14. For instance, with vital eight points $Ai$, they are intersection points of straight and curved lines as shown in Fig. 14, it is determined that the fire $F1$ is the nearest point with respect to point $A2$, so it is classified to group $A1A2$ and similarly the fire $F2$ belongs to group $A3A4$, and thus points $C1$ and $C2$ are defined. The remaining points $Ci$ are solved by the same way. Consequently, the mobile manipulator can complete its work with only a single movement around the fire test in the anti-clockwise direction.

Secondly, Algorithm I.2 as shown in Fig. 9 distributes the data determined in Algorithm I.1 to the chassis controller, driving the chassis to those desired positions $Ci$, then manipulator controller performs the end-effector to reach the flames $Fi$ and sprays the extinguishing agent. Some simple sub-functions in Algorithm I.1 and Algorithm I.2 are not presented in this paper.

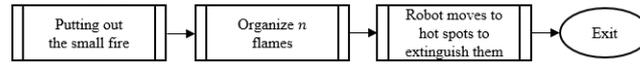

**Fig. 7.** Algorithm I for putting out the small fire

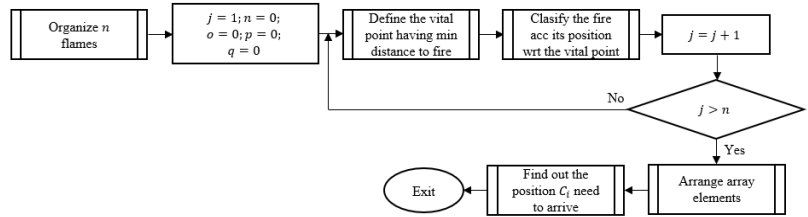

**Fig. 8.** Algorithm I.1 for putting out the small fire

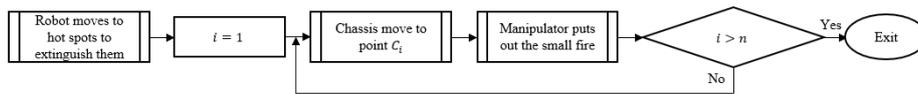

**Fig. 9.** Algorithm I.2 for making the mobile manipulator moves to the hot spots to extinguish them

## 2.2 Manipulator Controller

As depicted in Fig. 10, the angle reference of the manipulator $q_R = [\theta_{R1} \quad \theta_{R2} \quad \theta_{R3} \quad \theta_{R4}]^T$ is solved based on two inputs, chassis trajectory and desired end-effector trajectory. Then, the manipulator controller is designed to track the angle trajectory. As a result, the end-effector follows the designed plan.

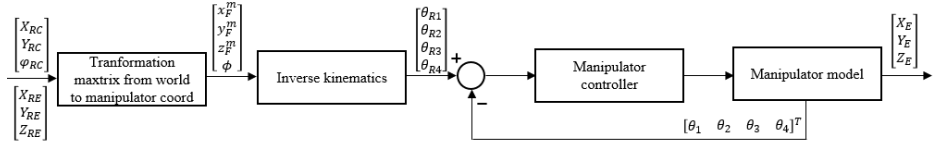

**Fig. 10.** Manipulator controller structure

The dynamic manipulator Eq. (3) with the control input $u$ is rewritten as follows:
$$M(q)\ddot{q} + C(q,\dot{q})\dot{q} + G(q) = u \quad (5)$$

Define the tracking error and the sliding surface as:
$$s = \dot{e} + \lambda e \quad (6)$$

where, $e = q_R - q$ and $\lambda > 0$, therefore, Eq. (6) is Hurwitz when $s = 0$.

The control law is defined as:
$$u = M(q)[\ddot{q}_R + \lambda \dot{e} + K sign(s)] + G(q) + C(q,\dot{q})\dot{q} \quad (7)$$

Choosing a Lyapunov candidate as $V = \frac{1}{2}s^T s$, by applying Barbalat's Lemma and Lyapunov stability theorem, it can be proved that $e \to 0$ as $t \to \infty$.

In the simulation result, the manipulator is assumed to be placed on the chassis moving at constant velocity $0.88 \, m/s$ and its end-effector velocity is $1.4 \, m/s$. The average tracking error of the end-effector are alternately $2.6 \, mm$, $1.8 \, mm$, $1.2 \, mm$ in axis $X, Y$ and $Z$. These errors are under the nozzle charge tolerance defined in key specification. The execution time of one edge in the entire trajectory is $2.956 \, s$, thus, the time spent for the whole trajectory in stage I is $11.824 \, s$. After that, the assumed time for stage II is $3 \, s$ to extinguish the small left flames. The combination of stages I and II consume $14.824 \, s$ which is less than the discharge time $15 \, s$ of the fire extinguisher for the class 20A fire test.

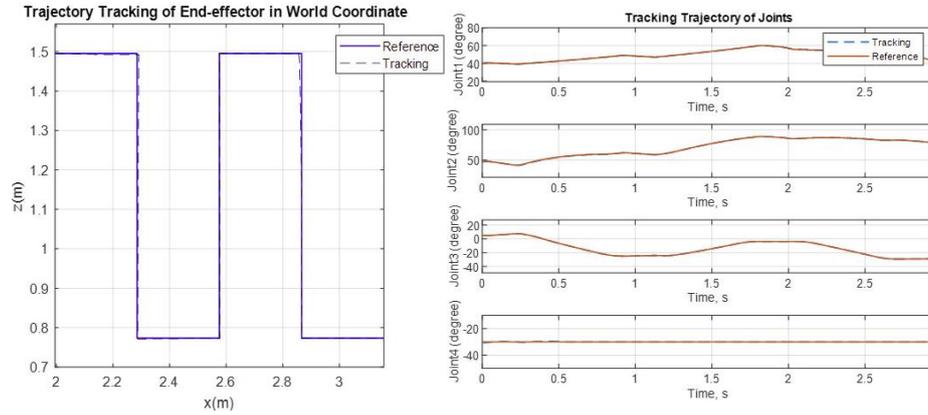

**Fig. 11.** The end-effector and joints of manipulator follow trajectory II, $v = 1.4 \, m/s$

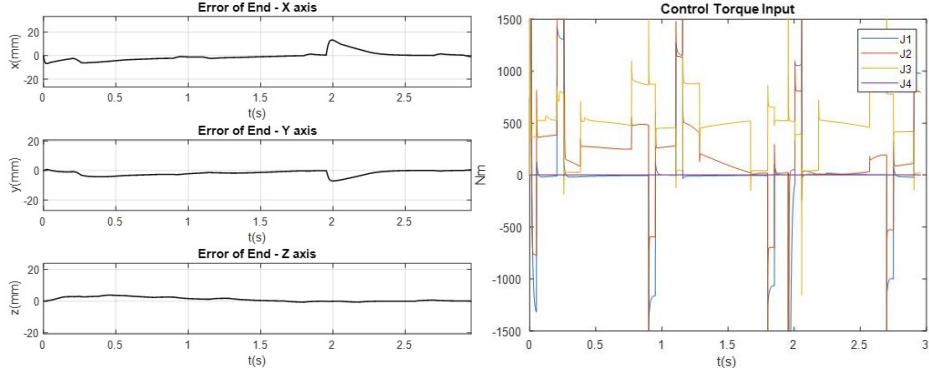

**Fig. 12.** Tracking error of the end-effector and joints and torques as input the manipulator, $v = 1.4\ m/s$

### 2.3 Chassis Controller

The chassis controller structure is described as in Fig. 13.

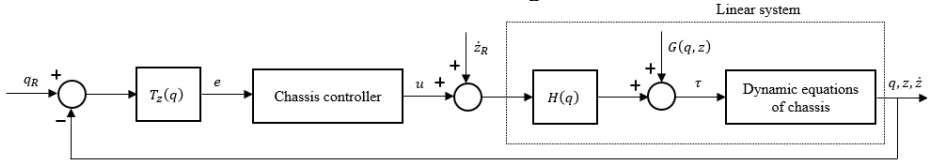

**Fig. 13.** Chassis controller structure

where, $q_R = [X_R\ \ Y_R\ \ \varphi_R]^T$ and $z_R = [v_R\ \ w_R]^T$ are the position and speed of trajectory; $q = [X_C\ \ Y_C\ \ \varphi_C]^T$ is the coordinates of the C.G. of the chassis in the global coordinate frame [5]; $z = [v\ \ w]^T$ is the speed of the C.G. of the chassis; $T_z(q)$ is the rotational matrix around z-axis [6], $e = [e_1\ \ e_2\ \ e_3]^T$ is the tracking error [6]; $\tau$ is the input torque; $H(q), G(q,z)$ are matrixes describing characteristics of the chassis [5].

To design the chassis controller, the sliding surface $S$ are defined as:

$$s = \begin{bmatrix} s_1 \\ s_2 \end{bmatrix} = \begin{bmatrix} \dot{e}_1 + k_1 e_1 + sgn(e_1)|\dot{e}_2 + k_2 e_2| \\ \dot{e}_3 + k_3 e_3 + \dot{e}_2 + k_2 e_2 \end{bmatrix} \quad (8)$$

where, $k_1, k_2$ and $k_3$ are positive values and $sgn(.)$ is the sign function.

When it reached the sliding surface, $s = 0$, the tracking error $e \to$ zero [6].

The fire extinguisher testing application does not demand moving through the unstructured environment, a sharp bend road at high speed or a slippery surface, so in the scope of this paper applying the proposed control law to the dynamic equations of chassis with the no-slipping assumption yields the linear system as following [6]:

$$\dot{z} - \dot{z}_R = u - f \quad (9)$$

where, $f = [f_1\ \ f_2]^T$; $|f_1| \leq f_{m1}, |f_2| \leq f_{m2}$, $f_{m1}$ and $f_{m2}$ are bounds of disturbance.

Here, the control input $u$ is chosen for stabilizing the sliding surface:

$$\begin{bmatrix}u_1\\u_2\end{bmatrix}=\begin{bmatrix}Q_1 & 0\\0 & Q_2\end{bmatrix}\begin{bmatrix}s_1\\s_2\end{bmatrix}+\begin{bmatrix}P_1 & 0\\0 & P_2\end{bmatrix}\begin{bmatrix}sgn(s_1)\\sgn(s_2)\end{bmatrix}+$$
$$\begin{bmatrix}\dot{e}_2 w+(e_2+l)\dot{w}-v_r\dot{e}_3\sin e_3\\0\end{bmatrix}+\begin{bmatrix}k_1\dot{e}_1+sgn(e_1)|\ddot{e}_2+k_2\dot{e}_2|\\k_3\dot{e}_3+\ddot{e}_2+k_2\dot{e}_2\end{bmatrix}+ \quad(10)$$
$$\begin{bmatrix}-\dot{v}_{d_x}\cos\varphi+v_{d_x}\dot{\varphi}\sin\varphi-\dot{v}_{d_y}\sin\varphi-v_{d_y}\dot{\varphi}\cos\varphi+\dot{w}_d e_2+w_d\dot{e}_2\\-\dot{w}_d\end{bmatrix}$$

After some calculations, differentiation of the sliding surface is given:
$$\dot{S}=-QS-P\,sgn(S)+f \quad(11)$$

Choosing a Lyapunov candidate as $V=\frac{1}{2}S^T S$. If constants $Q_i>0$, $P_i\geq f_{m_i}$ with $i=1,2$, then $\dot{V}<0$, and the control law stabilizes the sliding surface.

The simulation shows that the tracking errors go to approximate zero at about $0.4\,s$. Maximal $e_1, e_2$ and $e_3$ are $18.7\,mm, 0.06\,mm$ and $0.59°$ with the velocity of the chassis is $0.88\,m/s$ is acceptable for the fire extinguisher testing application.

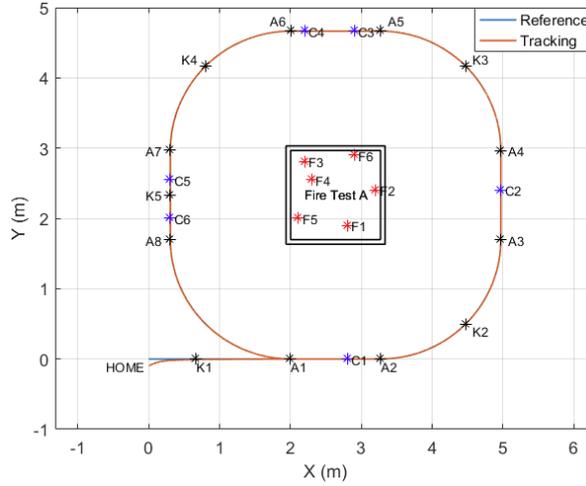

**Fig. 14.** Chassis follows trajectory II, $v=0.88\,m/s$

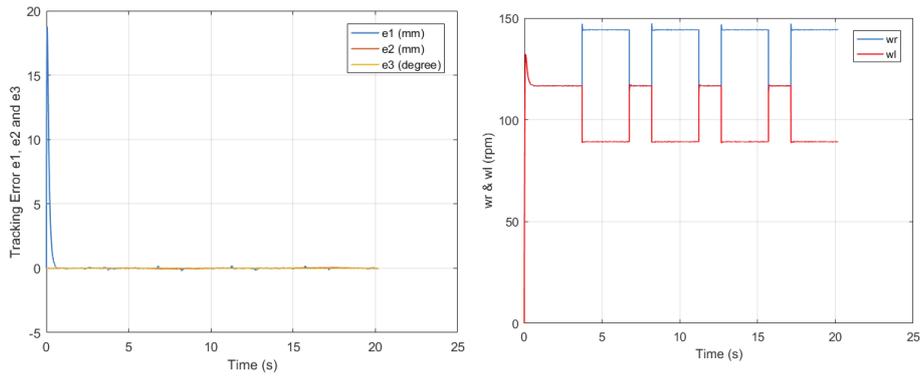

**Fig. 15.** Error tracking of chassis and Wheels velocity as input chassis, $v=0.88\,m/s$

## 3      Conclusion

Based on key specifications and conceptual design mentioned in Part I, the mobile manipulator design for fire extinguisher testing is presented in Part II.

On the one hand, from the simulation results, the designed mobile manipulator is proved to meet all requirements of the key specification. Specifically, recommending the algorithm enables the mobile manipulator to perform the fire extinguisher testing task continually, and the mobile manipulator completes the trajectory to extinguish the flame within the discharge time of the fire extinguisher. The tracking errors are acceptable compared to the tolerance criteria in key specification presented in Part I.

On the other hand, because these are first steps of developing the mobile manipulator for testing fire extinguisher, some proposals might not be the optimal solutions. Therefore, a lot of further potential work needs to be done to improve the design. For instance, optimizing dimension of linkages of the manipulator helps to avoid wasting the workspace; optimizing the trajectory and algorithms for putting out the fire tests allows the mobile manipulator to shorten the performing time; slip dynamics should be intergrated into the dynamic model to upgrade the tracked chassis capability of moving through rough terrain; the velocity of the chassis is modified so that when the manipulator finish trajectory I, the chassis must is at point $K2$.

## Acknowledgments

This research is funded by Vietnam National University Ho Chi Minh City (VNU-HCM) under grant number TX2022-20b-01. We acknowledge the support of time and facilities from National Key Laboratory of Digital Control and System Engineering (DCSELab), Ho Chi Minh City University of Technology (HCMUT), VNU-HCM for this study.